\documentclass[final]{cvpr}
\usepackage{times}
\usepackage{epsfig}
\usepackage{graphicx}
\usepackage{amsmath}
\usepackage{amssymb}

\usepackage{array,booktabs,calc}
\usepackage{placeins}
\usepackage[inline, shortlabels]{enumitem}
\usepackage{graphicx}
\usepackage{multirow}
\usepackage{wrapfig}
\usepackage{paralist}
\usepackage[dvipsnames]{xcolor}
\usepackage{inconsolata}
\usepackage{gensymb}
\usepackage{caption} 
\usepackage{bbm}  %
\usepackage{xcolor}
\usepackage{floatrow}
\usepackage{nicefrac}
\usepackage{subfig}
\usepackage{stfloats}
\usepackage[pagebackref, pageanchor=true, plainpages=false, pdfpagelabels, bookmarks, bookmarksnumbered]{hyperref}

\setlist{itemjoin ={,\enspace},itemjoin* = { and\enspace}}
\definecolor{citecolor}{HTML}{0071bc}
\hypersetup{
  breaklinks   = true,
  colorlinks   = true, %
  urlcolor     = RubineRed, %
  linkcolor    = RubineRed, %
  citecolor    = citecolor %
}
\def\cvprPaperID{3434} %
\def\confYear{CVPR 2021}

\newcommand{\PAR}[1]{\vskip4pt \noindent{\bf #1~}}
\newcommand{\minus}{\scalebox{0.75}[1.0]{$-$}}

\renewcommand{\b}[1]{\textbf{#1}}
\newcommand{\0}{\phantom{0}}
\newcommand\blfootnote[1]{%
  \begingroup
  \renewcommand\thefootnote{}\footnote{#1}%
  \addtocounter{footnote}{-1}%
  \endgroup
}

\newcommand{\shortname}{LoFTR\xspace}

\begin{document}

\title{LoFTR: Detector-Free Local Feature Matching with Transformers}

\author{
    Jiaming Sun$^{1,2*}$ 
    \quad Zehong Shen$^{1*}$ 
    \quad Yuang Wang$^{1*}$ 
    \quad Hujun Bao$^{1}$ 
    \quad Xiaowei Zhou$^{1\dagger}$\\[1.5mm]
    $^1$Zhejiang University \quad 
    $^2$SenseTime Research\quad
}

\maketitle

\begin{abstract}
  \textit{
We present a novel method for local image feature matching.  Instead of performing image feature detection, description, and matching sequentially,  we propose to first establish pixel-wise dense matches at a coarse level and later refine the good matches at a fine level. In contrast to dense methods that use a cost volume to search correspondences,  we use self and cross attention layers in Transformer to obtain feature descriptors that are conditioned on both images. The global receptive field provided by Transformer enables our method to produce dense matches in low-texture areas,  where feature detectors usually struggle to produce repeatable interest points. The experiments on indoor and outdoor datasets show that LoFTR outperforms state-of-the-art methods by a large margin.  
LoFTR also ranks first on two public benchmarks of visual localization among the published methods.
Code is available at our project page: \url{https://zju3dv.github.io/loftr/}.
}

\end{abstract}

\blfootnote{$^*$The first three authors contributed equally. The authors are affiliated with the State Key Lab of CAD\&CG and ZJU-SenseTime Joint Lab of 3D Vision. $^\dagger$Corresponding author: Xiaowei Zhou.}

\section{Introduction}\label{sec:intro}
Local feature matching between images is the cornerstone of many 3D computer vision tasks,
including structure from motion (SfM), simultaneous localization and mapping (SLAM), visual localization, etc.
Given two images to be matched, 
most existing matching methods consist of three separate phases: feature detection, feature description, and feature matching.
In the detection phase, salient points like corners are first detected as interest points from each image.
Local descriptors are then extracted around neighborhood regions of these interest points.
The feature detection and description phases produce two sets of interest points with descriptors, the point-to-point correspondences of which are later found by nearest neighbor search or more sophisticated matching algorithms.  

\begin{figure}[tb]
    \centering
    \includegraphics[width=0.95\linewidth]{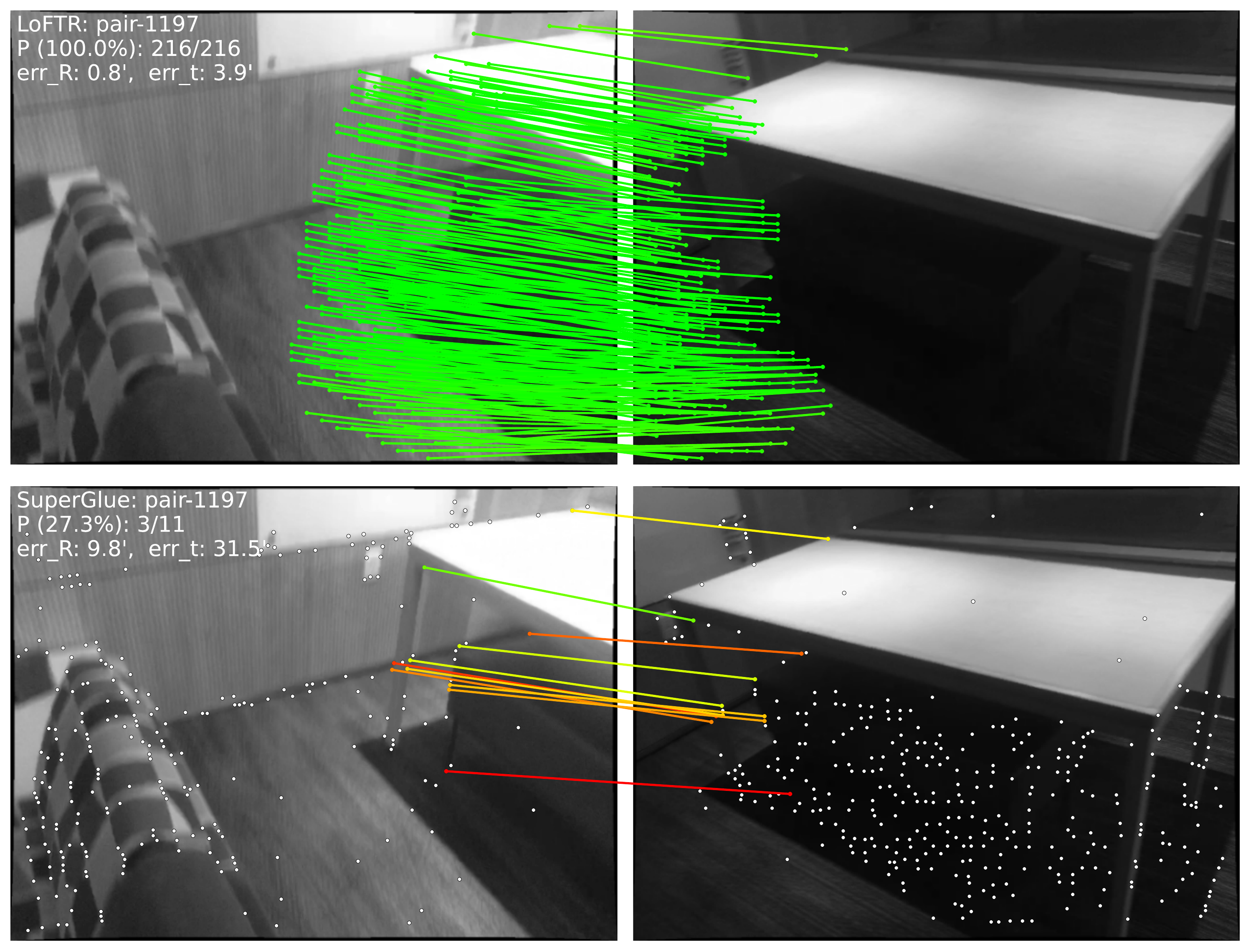}
    \caption[Comparison between the proposed method \shortname and the detector-based method SuperGlue~\cite{sarlinSuperGlueLearningFeature2019}.]{
        \textbf{Comparison between the proposed method \shortname and the detector-based method SuperGlue~\cite{sarlinSuperGlueLearningFeature2019}.}
        This example demonstrates that \shortname is capable of finding correspondences on the texture-less wall and the floor with repetitive patterns, where detector-based methods struggle to find repeatable interest points.\footnotemark
    }
    \label{fig:teaser}
\end{figure}

The use of a feature detector reduces the search space of matching, and the resulted sparse correspondences are sufficient for most tasks, e.g., camera pose estimation.
However, a feature detector may fail to extract enough interest points that are repeatable between images due to various factors such as poor texture, repetitive patterns, viewpoint change, illumination variation, and motion blur.
This issue is especially prominent in indoor environments, where low-texture regions or repetitive patterns sometimes occupy most areas in the field of view. Fig. \ref{fig:teaser} shows an example. Without repeatable interest points, it is impossible to find correct correspondences even with perfect descriptors.
\footnotetext{Only the inlier matches after RANSAC are shown. The green color indicates a match with epipolar error smaller than $5 \times 10^{-4}$ (in the normalized image coordinates).}

Several recent works~\cite{roccoNeighbourhoodConsensusNetworks2018,roccoEfficientNeighbourhoodConsensus2020,liDualResolutionCorrespondenceNetworks2020} have attempted to remedy this problem
by establishing pixel-wise dense matches.
Matches with high confidence scores can be selected from the dense matches, and thus feature detection is avoided. 
However, the dense features extracted by convolutional neural networks (CNNs) in these works have limited receptive field which may not distinguish indistinctive regions. 
Instead, humans find correspondences in these indistinctive regions not only based on the \textit{local} neighborhood, but with a larger \textit{global} context.
For example, low-texture regions in Fig. \ref{fig:teaser} can be distinguished according to their relative positions to the edges.
This observation tells us that a large \textit{receptive field} in the feature extraction network is crucial.

Motivated by the above observations, we propose Local Feature TRansformer (\shortname), a novel detector-free approach to local feature matching.
Inspired by seminal work SuperGlue~\cite{sarlinSuperGlueLearningFeature2019},
we use Transformer~\cite{vaswaniAttentionAllYou2017} with self and cross attention layers to process (transform) the dense local features extracted from the convolutional backbone.
Dense matches are first extracted between the two sets of transformed features at a low feature resolution ($\nicefrac{1}{8}$ of the image dimension).
Matches with high confidence are selected from these dense matches and later refined to a sub-pixel level with a correlation-based approach.
The global receptive field and positional encoding of Transformer enable the transformed feature representations to be context- and position-dependent.
By interleaving the self and cross attention layers multiple times, LoFTR learns the densely-arranged globally-consented matching priors exhibited in the ground-truth matches.
A linear transformer is also adopted to reduce the computational complexity to a manageable level.

We evaluate the proposed method on several image matching and camera pose estimation tasks with indoor and outdoor datasets. 
The experiments show that \shortname outperforms detector-based and detector-free feature matching baselines by a large margin.
\shortname also achieves state-of-the-art performance and ranks first among the published methods on two public benchmarks of visual localization.
Compared to detector-based baseline methods, 
\shortname can produce high-quality matches even in indistinctive regions with low-textures, motion blur, or repetitive patterns.

\section{Related Work}\label{sec:related}
\PAR{Detector-based Local Feature Matching.}
Detector-based methods have been the dominant approach for local feature matching.
Before the age of deep learning, many renowned works in the traditional hand-crafted local features have achieved good performances.
SIFT~\cite{loweDistinctiveImageFeatures2004} and ORB~\cite{rubleeORBEfficientAlternative2011} are arguably the most successful hand-crafted local features and are widely adopted in many 3D computer vision tasks.
The performance on large viewpoint and illumination changes of local features can be significantly improved with learning-based methods.
Notably, LIFT~\cite{yiLIFTLearnedInvariant2016} and MagicPoint~\cite{detoneGeometricDeepSLAM2017} are among the first successful learning-based local features.
They adopt the detector-based design in hand-crafted methods and achieve good performance.
SuperPoint~\cite{detoneSuperPointSelfSupervisedInterest2018} builds upon MagicPoint and proposes a self-supervised training method through homographic adaptation.
Many learning-based local features along this line~\cite{revaudR2D2RepeatableReliable,dusmanuD2NetTrainableCNN2019,liuGIFTLearningTransformationInvariant2019,luoASLFeatLearningLocal2020,tyszkiewiczDISKLearningLocal2020}
also adopt the detector-based design.

The above-mentioned local features use the nearest neighbor search to find matches between the extracted interest points.
Recently, SuperGlue~\cite{sarlinSuperGlueLearningFeature2019} proposes a learning-based approach for local feature matching.
SuperGlue accepts two sets of interest points with their descriptors as input and learns their matches with a graph neural network (GNN), which is a general form of Transformers~\cite{joshi2020transformers}.
Since the priors in feature matching can be learned with a data-driven approach, SuperGlue achieves impressive performance and sets the new state of the art in local feature matching.
However, being a detector-dependent method, it has the fundamental drawback of being unable to detect repeatable interest points in indistinctive regions.
The attention range in SuperGlue is also limited to the detected interest points only.
Our work is inspired by SuperGlue in terms of using self and cross attention in GNN for message passing between two sets of descriptors, 
but we propose a detector-free design to avoid the drawbacks of feature detectors.
We also use an efficient variant of the attention layers in Transformer to reduce the computation costs.

\PAR{Detector-free Local Feature Matching.}
Detector-free methods remove the feature detector phase and directly produce dense descriptors or dense feature matches.
The idea of dense features matching dates back to SIFT Flow~\cite{liuSIFTFlowDense}.
\cite{choyUniversalCorrespondenceNetwork2016,schmidtSelfSupervisedVisualDescriptor2017} are the first learning-based approaches to learn pixel-wise feature descriptors with the contrastive loss.
Similar to the detector-based methods, the nearest neighbor search is usually used as a post-processing step to match the dense descriptors.
NCNet~\cite{roccoNeighbourhoodConsensusNetworks2018} proposed a different approach by directly learning the dense correspondences in an end-to-end manner.
It constructs 4D cost volumes to enumerate all the possible matches between the images and uses 4D convolutions to regularize the cost volume and enforce neighborhood consensus among all the matches.
Sparse NCNet~\cite{roccoEfficientNeighbourhoodConsensus2020} improves upon NCNet and makes it more efficient with sparse convolutions.
Concurrently with our work, DRC-Net~\cite{liDualResolutionCorrespondenceNetworks2020} follows this line of work and proposes a coarse-to-fine approach to produce dense matches with higher accuracy.
Although all the possible matches are considered in the 4D cost volume, the receptive field of 4D convolution is still limited to each matches' neighborhood area.
Apart from neighborhood consensus, 
our work focuses on achieving global consensus between matches with the help of the global receptive field in Transformers, which is not exploited in NCNet and its follow-up works.
\cite{liu2020extremely} proposes a dense matching pipeline for SfM with endoscopy videos.
The recent line of research~\cite{GLUNet_Truong_2020,GOCor_Truong_2020,Truong2021LearningAD,jiang2021cotr} that focuses on bridging the task of local feature matching and optical flow estimation, is also related to our work.

\begin{figure*}[!t]
    \vspace{-1.2cm}
    \centering
    \includegraphics[width=0.98\linewidth]{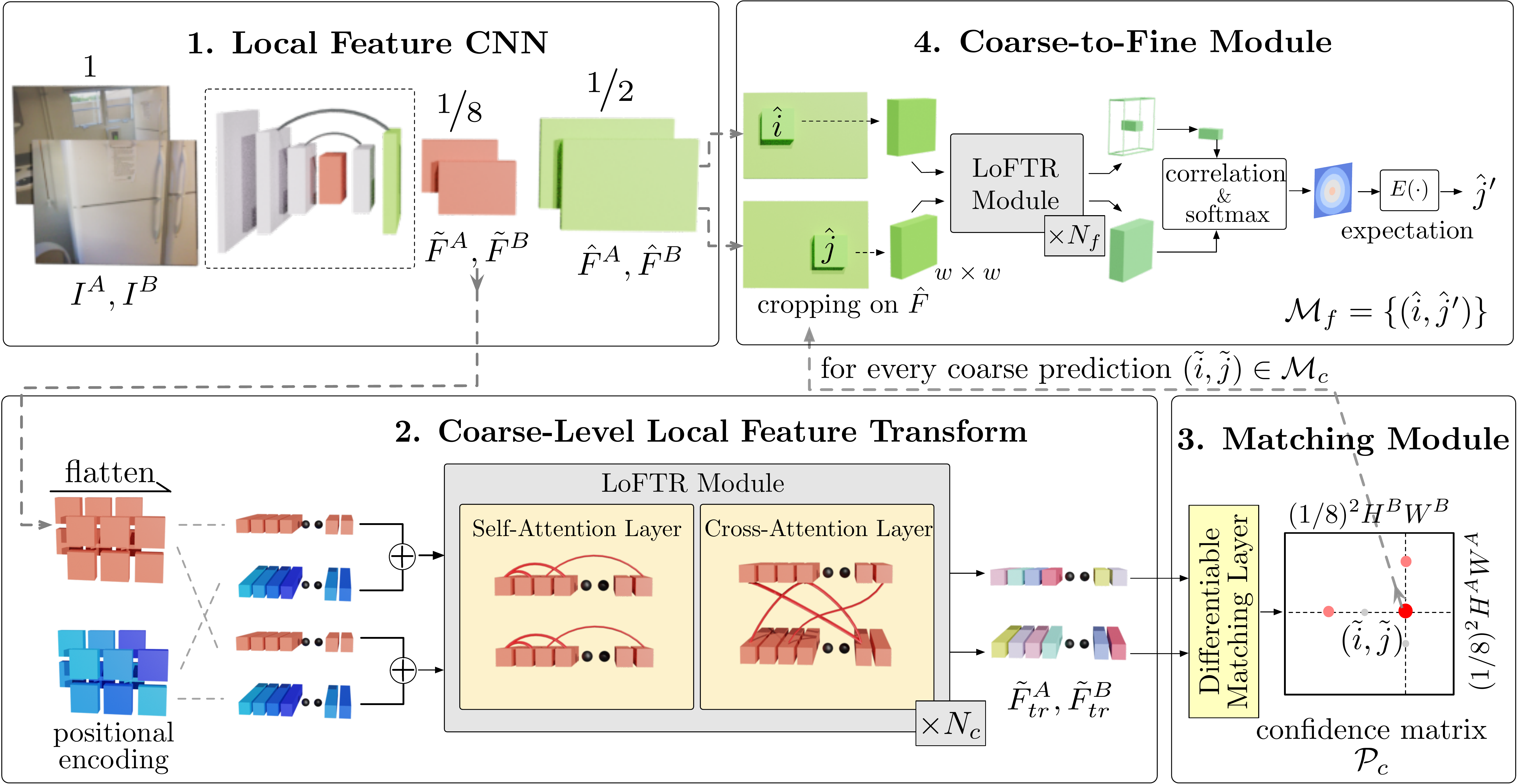}%
    \caption{\textbf{Overview of the proposed method.}
        LoFTR has four components:
        \textbf{1.} A local feature CNN extracts
        the coarse-level feature maps 
        $\tilde{F}^A$ and $\tilde{F}^B$,
        together with the fine-level feature maps 
        $\hat{F}^A$ and $\hat{F}^B$
        from the image pair $I^A$ and $I^B$ (Section \ref{sec:local}).
        \textbf{2.}  The coarse feature maps are flattened to 1-D vectors and added with the
        positional encoding.
        The added features are then processed by the Local Feature TRansformer (LoFTR) module, which has $N_c$ self-attention and cross-attention layers (Section \ref{sec:loftr}).
        \textbf{3.} A differentiable matching layer is used to match the transformed features, which ends up with a confidence matrix $\mathcal{P}_c$. 
        The matches in $\mathcal{P}_c$ are selected according to the confidence threshold and mutual-nearest-neighbor criteria, 
        yielding the coarse-level match prediction $\mathcal{M}_c$ (Section \ref{sec:sinkhorn}).
        \textbf{4.} For every selected coarse prediction $(\tilde{i}, \tilde{j}) \in \mathcal{M}_c$,
        a local window with size $w\times w$ is cropped from the fine-level feature map.
        Coarse matches will be refined within this local window to a sub-pixel level as the final match prediction $\mathcal{M}_f$
        (Section \ref{sec:coarse_to_fine}).
    }
    \label{fig:overview}
\end{figure*}

\PAR{Transformers in Vision Related Tasks.}
Transformer~\cite{vaswaniAttentionAllYou2017} has become the \textit{de facto} standard for sequence modeling in natural language processing (NLP) due to their simplicity and computation efficiency.
Recently, Transformers are also getting more attention in computer vision tasks, such as 
image classification~\cite{anonymousImageWorth16x162020}, object detection~\cite{carionEndtoEndObjectDetection2020} and semantic segmentation~\cite{wangAxialDeepLabStandAloneAxialAttention2020}.
Concurrently with our work, \cite{liRevisitingStereoDepth2020} proposes to use Transformer for disparity estimation.
The computation cost of the vanilla Transformer grows quadratically as the length of input sequences due to the multiplication between query and key vectors.
Many efficient variants~\cite{tayEfficientTransformersSurvey2020,kitaevReformerEfficientTransformer2020,katharopoulosTransformersAreRNNs2020,choromanskiRethinkingAttentionPerformers2020} are proposed recently in the context of processing long language sequences. 
Since no assumption of the input data is made in these works, they are also well suited for processing images.

\section{Methods}\label{sec:method}
Given the image pair $I^A$ and $I^B$, the existing local feature matching methods use a feature detector to extract interest points.
We propose to tackle the repeatability issue of feature detectors with a detector-free design.
An overview of the proposed method \shortname is presented in Fig.~\ref{fig:overview}.

\subsection{Local Feature Extraction}\label{sec:local}
We use a standard convolutional architecture with FPN~\cite{Lin2017FeaturePN} (denoted as the local feature CNN) to extract multi-level features from both images.
We use $\tilde{F}^A$ and $\tilde{F}^B$ to denote the coarse-level features at $\nicefrac{1}{8}$ of the original image dimension,
and $\hat{F}^A$ and $\hat{F}^B$ the fine-level features at $\nicefrac{1}{2}$ of the original image dimension.

Convolutional Neural Networks (CNNs) possess the inductive bias of translation equivariance and locality, which are well suited for \textit{local} feature extraction.
The downsampling introduced by the CNN also reduces the input length of the \shortname module, which is crucial to ensure a manageable computation cost.

\subsection{Local Feature Transformer (LoFTR) Module}\label{sec:loftr}
After the local feature extraction, 
$\tilde{F}^A$ and $\tilde{F}^B$ are 
passed through the LoFTR module to extract position and context dependent local features.
Intuitively, the LoFTR module transforms the features into feature representations that are easy to match.
We denote the transformed features as $\tilde{F}^A_{tr}$ and $\tilde{F}^B_{tr}$.
\begin{figure}[ht!]
    \vspace{-1.2cm}
    \centering
    \includegraphics[width=0.9\linewidth]{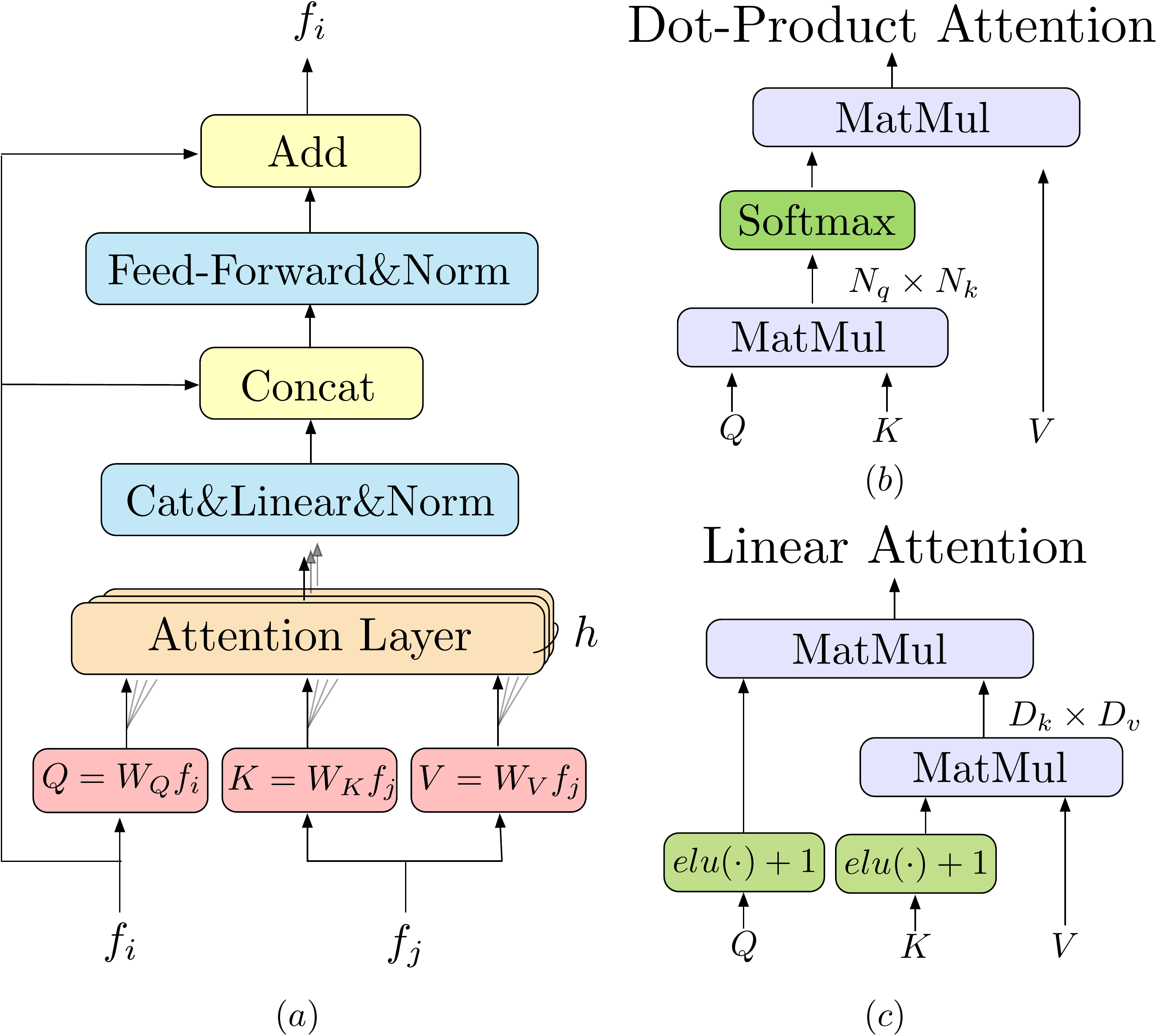}%
    \caption{\textbf{Encoder layer and attention layer in \shortname.} 
    (a) Transformer encoder layer. $h$ represents the multiple heads of attention.
    (b) Vanilla dot-product attention with $O(N^2)$ complexity. 
    (c) Linear attention layer with $O(N)$ complexity. 
    The scale factor is omitted for simplicity.}
    \label{fig:tr_module}%
\end{figure}

\PAR{Preliminaries: Transformer~\cite{vaswaniAttentionAllYou2017}.}
We first briefly introduce the Transformer here as background.
A Transformer encoder is composed of sequentially connected encoder layers.
Fig.~\ref{fig:tr_module}(a) shows the architecture of an encoder layer.

The key element in the encoder layer is the attention layer.
The input vectors for an attention layer are conventionally named query, key, and value.
Analogous to information retrieval, the query vector $Q$ retrieves information from the value vector $V$, 
according to the attention weight computed from the dot product of $Q$ and the key vector $K$ corresponding to each value $V$.
The computation graph of the attention layer is presented in Fig.~\ref{fig:tr_module}(b).
Formally, the attention layer is denoted as:
\begin{equation*}
    \text{Attention}(Q, K, V) = \operatorname{softmax}(QK^T)V.
    \label{eq:attn}
\end{equation*}
Intuitively, the attention operation selects the relevant information by measuring the similarity between the query element and each key element.  
The output vector is the sum of the value vectors weighted by the similarity scores.
As a result, the relevant information is extracted from the value vector if the similarity is high.
This process is also called ``message passing'' in Graph Neural Network.

\begin{figure}[ht!]
    \vspace{-1.2cm}
    \centering
    \includegraphics[width=\linewidth]{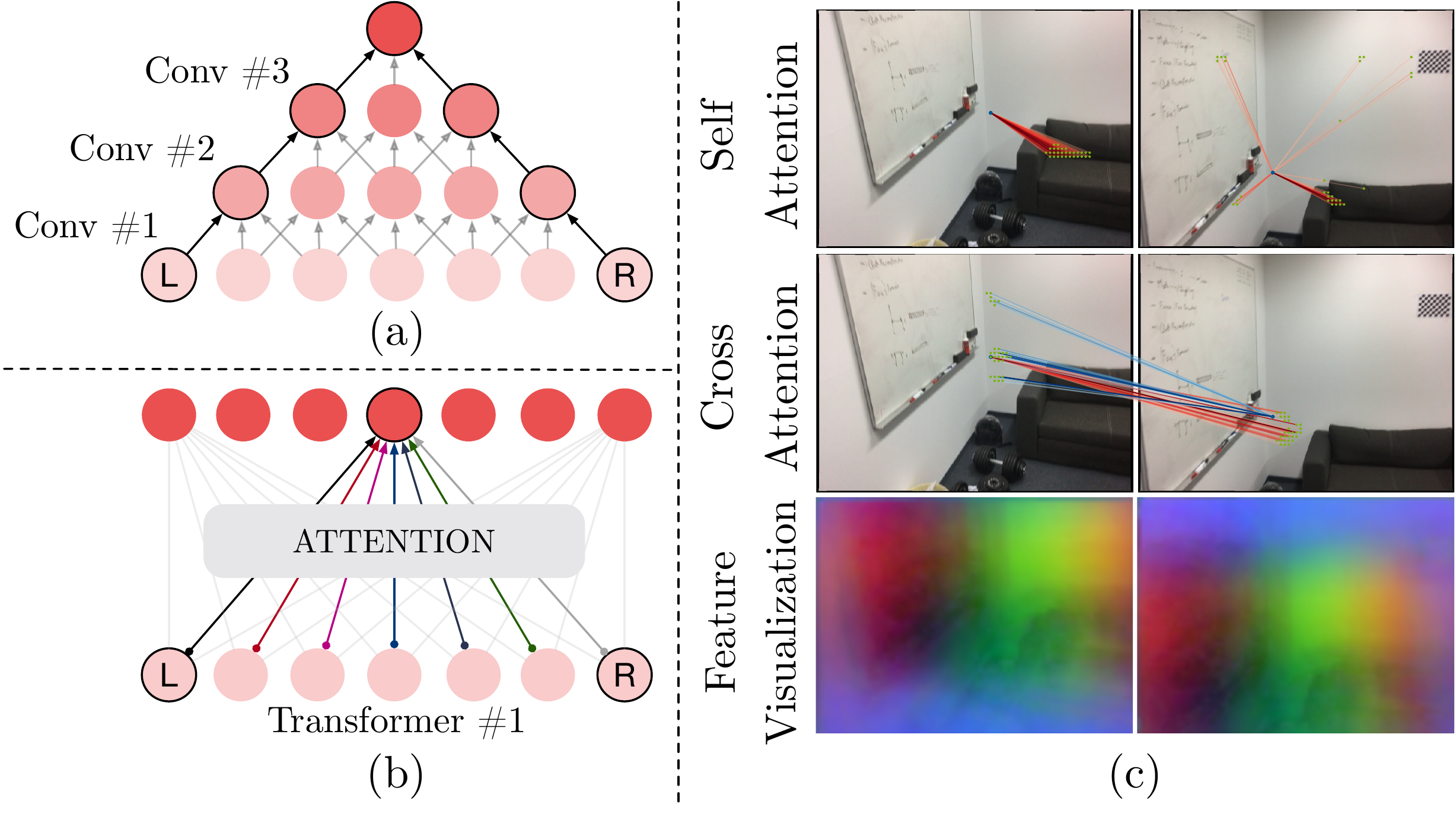}%
    \caption{Illustration of the receptive field of \textbf{(a)} Convolutions and \textbf{(b)} Transformers.
    Assume that the objective is to establish a connection between the L and R elements to extract their joint feature representation. 
    Due to the local-connectivity of convolutions, many convolution layers need to be stacked together in order to achieve this connection.
    The global receptive field of Transformers enables this connection to be established through only one attention layer.
    \textbf{(c)} Visualization of the attention weights and transformed dense features.
    We use PCA to reduce the dimension of the transformed features $\tilde{F}^A_{tr}$ and $\tilde{F}^B_{tr}$ and visualize the results with RGB color.
    Zoom in for details.
    }
    \label{fig:fig-fig}%
\end{figure}

\PAR{Linear Transformer.}\label{sec:lin-tr}
Denoting the length of $Q$ and $K$ as $N$ and their feature dimension as $D$, 
the dot product between $Q$ and $K$ in the Transformer introduces computation cost that grows quadratically ($O(N^2)$) with the length of the input sequence.
Directly applying the vanilla version of Transformer in the context of local feature matching is impractical even when the input length is reduced by the local feature CNN. 
To remedy this problem, we propose to use an efficient variant of the vanilla attention layer in Transformer.
Linear Transformer~\cite{katharopoulosTransformersAreRNNs2020} 
proposes to reduce the computation complexity of Transformer to $O(N)$ by substituting the exponential kernel used in the original attention layer with an alternative kernel function
$\operatorname{sim}(Q, K) = \phi(Q) \cdot \phi(K)^T, \text{where } \phi(\cdot) = \operatorname{elu}(\cdot) + 1$.
This operation is illustrated by the computation graph in Fig.~\ref{fig:tr_module}(c).
Utilizing the associativity property of matrix products, the multiplication between $\phi(K)^T$ and $V$ can be carried out first. 
Since $D \ll N$, the computation cost is reduced to $O(N)$.

\PAR{Positional Encoding.}
We use the 2D extension of the standard positional encoding in Transformers following DETR~\cite{carionEndtoEndObjectDetection2020}. Different from DETR, we only add them to the backbone output once.
We leave the formal definition of the positional encoding in the supplementary material.
Intuitively, the positional encoding gives each element unique position information in the sinusoidal format.
By adding the position encoding to $\tilde{F}^A$ and $\tilde{F}^B$, the transformed features will become position-dependent, which is crucial to the ability of LoFTR to produce matches in indistinctive regions.
As shown in the bottom row of Fig.~\ref{fig:fig-fig}(c), although the input RGB color is homogeneous on the white walls, 
the transformed features $\tilde{F}^A_{tr}$ and $\tilde{F}^B_{tr}$ are \textit{unique} for each position demonstrated by the smooth color gradients.
More visualizations are provided in Fig.~\ref{fig:qua-attention}.

\PAR{Self-attention and Cross-attention Layers.}
For self-attention layers, the input features $f_i$ and $f_j$ (shown in Fig.~\ref{fig:tr_module}) are the same (either $\tilde{F}^A$ or $\tilde{F}^B$).
For cross-attention layers, the input features $f_i$ and $f_j$ are either ($\tilde{F}^A$ and $\tilde{F}^B$) or ($\tilde{F}^B$ and $\tilde{F}^A$) depending on the direction of cross-attention.
Following \cite{sarlinSuperGlueLearningFeature2019}, we interleave the self and cross attention layers in the LoFTR module by $N_c$ times.
The attention weights of the self and cross attention layers in LoFTR are visualized in the first two rows of Fig.~\ref{fig:fig-fig}(c).

\subsection{Establishing Coarse-level Matches}\label{sec:sinkhorn}

Two types of differentiable matching layers can be applied in \shortname, either with an optimal transport (OT) layer as in ~\cite{sarlinSuperGlueLearningFeature2019} or with a dual-softmax operator~\cite{roccoNeighbourhoodConsensusNetworks2018,tyszkiewiczDISKLearningLocal2020}.
The score matrix $\mathcal{S}$ between the transformed features is first calculated by $\mathcal{S}\left(i, j\right) = \frac{1}{\tau} \cdot \langle\tilde{F}_{t r}^{A}(i), \tilde{F}_{t r}^{B}(j)\rangle$.
When matching with OT, $\minus \mathcal{S}$
can be used as the cost matrix of the partial assignment problem as in~\cite{sarlinSuperGlueLearningFeature2019}. 
We can also apply softmax on both dimensions (referred to as dual-softmax in the following) of $\mathcal{S}$ to obtain the probability of soft mutual nearest neighbor matching. 
Formally, when using dual-softmax, the matching probability $\mathcal{P}_{c}$ is obtained by:
\begin{equation*}
    \mathcal{P}_{c}(i, j) = \operatorname{softmax}\left(\mathcal{S}\left(i, \cdot \right)\right)_j \cdot \operatorname{softmax}\left(\mathcal{S}\left(\cdot, j\right)\right)_i.
    \label{eq:dual-softmax}
\end{equation*}

\PAR{Match Selection.}
Based on the confidence matrix $\mathcal{P}_{c}$, 
we select matches with confidence higher than a threshold of $\theta_{c}$, 
and further enforce the mutual nearest neighbor (MNN) criteria, which filters possible outlier coarse matches.
We denote the coarse-level match predictions as:
\begin{equation*}
    \mathcal{M}_{c} = \{\left(\tilde{i},\tilde{j}\right) \mid \forall\left(\tilde{i},\tilde{j}\right) \in \operatorname{MNN}\left(\mathcal{P}_{c}\right),\ \mathcal{P}_{c}\left(\tilde{i},\tilde{j}\right) \geq \theta_{c}\}.
    \label{eq:coarse-matches}
\end{equation*}

\subsection{Coarse-to-Fine Module}\label{sec:coarse_to_fine}
After establishing coarse matches, these matches are refined to the original image resolution with the coarse-to-fine module.
Inspired by~\cite{wangLearningFeatureDescriptors2020}, we use a correlation-based approach for this purpose.
For every coarse match $(\tilde{i}, \tilde{j})$, we first locate its position $(\hat{i}, \hat{j})$ at fine-level feature maps $\hat F^A$ and $\hat F^B$, and then crop two sets of local windows of size $w\times w$. 
A smaller LoFTR module then transforms the cropped features within each window by $N_f$ times, yielding two transformed local feature maps $\hat{F}^A_{tr}(\hat{i})$ and ${\hat{F}^B_{tr}(\hat{j})}$ centered at $\hat{i}$ and $\hat{j}$, respectively.
Then, we correlate the center vector of $\hat{F}^A_{tr}(\hat{i})$ with all vectors in ${\hat{F}^B_{tr}(\hat{j})}$ and thus produce a heatmap that represents the matching probability of each pixel in the neighborhood of $\hat{j}$ with $\hat{i}$.
By computing expectation over the probability distribution, we get the final position $\hat{j}'$ with sub-pixel accuracy on $I^B$.
Gathering all the matches $\{(\hat{i}, \hat{j}^\prime)\}$ produces the final fine-level matches $\mathcal{M}_f$.

\subsection{Supervision}
The final loss consists of the losses for the coarse-level and the fine-level: %
$\mathcal{L} = \mathcal{L}_{c} + \mathcal{L}_{f}$.

\PAR{Coarse-level Supervision.} 
The loss function for the coarse-level is the negative log-likelihood loss over the confidence matrix $\mathcal{P}_{c}$ returned by either the optimal transport layer or the dual-softmax operator.
We follow SuperGlue~\cite{sarlinSuperGlueLearningFeature2019} to use camera poses and depth maps to compute the ground-truth labels for the confidence matrix during training.
We define the ground-truth coarse matches $\mathcal{M}_{c}^{gt}$ as the mutual nearest neighbors of the two sets of \nicefrac{1}{8}-resolution grids.
The distance between two grids is measured by the re-projection distance of their central locations. 
More details are provided in the supplementary.
With the optimal transport layer, we use the same loss formulation as in \cite{sarlinSuperGlueLearningFeature2019}.
When using dual-softmax for matching, we minimize the negative log-likelihood loss over the grids in $\mathcal{M}_{c}^{gt}$:
\begin{equation*}
    \mathcal{L}_{c} = - \frac{1}{\lvert\mathcal{M}_{c}^{gt}\rvert}
    \sum_{(\tilde{i}, \tilde{j}) \in \mathcal{M}_{c}^{gt}} \log \mathcal{P}_{c}\left(\tilde{i}, \tilde{j}\right).
    \label{eq:loss_coarse}
\end{equation*}

\PAR{Fine-level Supervision.} We use the $\ell_2$ loss for fine-level refinement.
Following~\cite{wangLearningFeatureDescriptors2020}, for each query point $\hat{i}$, we also measure its uncertainty by calculating the total variance $\sigma^2(\hat{i})$ of the corresponding heatmap. 
The target is to optimize the refined position that has low uncertainty, resulting in the final weighted loss function:
\begin{equation*}
    \mathcal{L}_{f}=\frac{1}{\lvert\mathcal{M}_{f} \rvert}\sum_{(\hat{i}, \hat{j}^{\prime}) \in \mathcal{M}_{f}} \frac{1}{\sigma^{2}(\hat{i})}\left\|\hat{j}^{\prime}-\hat{j}^{\prime}_{g t}\right\|_{2},
    \label{eq:loss_fine}
\end{equation*}
in which $\hat{j}^{\prime}_{gt}$ is calculated by 
warping each $\hat{i}$ from $\hat{F}^A_{tr}(\hat{i})$ to $\hat{F}^B_{tr}(\hat{j})$ 
with the ground-truth camera pose and depth.
We ignore ($\hat{i}$, $\hat{j}^\prime$) if the warped location of $\hat{i}$ falls out of the local window of $\hat{F}^B_{tr}(\hat{j})$ when calculating $\mathcal{L}_{f}$. The gradient is not backpropagated through $\sigma^{2}(\hat{i})$ during training.

\subsection{Implementation Details}
We train the indoor model of LoFTR on the ScanNet~\cite{daiScanNetRichlyannotated3D2017} dataset and the outdoor model on the MegaDepth~\cite{liMegaDepthLearningSingleView2018} following~\cite{sarlinSuperGlueLearningFeature2019}. 
On ScanNet, the model is trained
using Adam with an initial learning rate of $1\times10^{-3}$ and a batch size of 64.
It converges after 24 hours of training on 64 GTX 1080Ti GPUs. 
The local feature CNN uses a modified version of ResNet-18~\cite{heDeepResidualLearning2016} as the backbone.
The entire model is trained end-to-end with randomly initialized weights.
$N_c$ is set to 4 and $N_f$ is 1.
$\theta_{c}$ is chosen to 0.2.
Window size $w$ is equal to 5.
$\tilde{F}^A_{tr}$ and $\tilde{F}^B_{tr}$ are upsampled and concatenated with $\hat{F}^A$ and $\hat{F}^B$ before passing through the fine-level LoFTR in the implementation.
The full model with dual-softmax matching runs at 116 ms for a 640$\times$480 image pair on an RTX 2080Ti. 
Under the optimal transport setup, we use three sinkhorn iterations, and the model runs at 130 ms.
We refer readers to the supplementary material for more details of training and timing analyses.

\section{Experiments}\label{sec:exp}

\subsection{Homography Estimation}
In the first experiment, we evaluate LoFTR on the widely adopted HPatches dataset~\cite{balntasHPatchesBenchmarkEvaluation2017} for homography estimation. 
HPatches contains 52 sequences under significant illumination changes and 56 sequences that exhibit large variation in viewpoints.

\PAR{Evaluation protocol.} In every test sequence, one reference image is paired with the rest five images. 
All images are resized with shorter dimensions equal to 480. 
For each image pair, we extract a set of matches with LoFTR trained on MegaDepth~\cite{liMegaDepthLearningSingleView2018}.
We use OpenCV to compute the homography estimation with RANSAC as the robust estimator. 
To make a fair comparison to methods that produce different numbers of matches,
we compute the corner error between the images warped with the estimated $\hat{\mathcal{H}}$ and the ground-truth $\mathcal{H}$ as a correctness identifier as in~\cite{detoneSuperPointSelfSupervisedInterest2018}. 
Following ~\cite{sarlinSuperGlueLearningFeature2019}, we report the area under the cumulative curve (AUC) of the corner error up to threshold values of 3, 5, and 10 pixels, respectively. 
We report the results of LoFTR with a maximum of 1K output matches.

\PAR{Baseline methods.} We compare LoFTR with three categories of methods:
1) detector-based local features including  R2D2~\cite{revaudR2D2RepeatableReliable}, D2Net~\cite{dusmanuD2NetTrainableCNN2019},
and DISK~\cite{tyszkiewiczDISKLearningLocal2020},
2) a detector-based local feature matcher, i.e., SuperGlue~\cite{sarlinSuperGlueLearningFeature2019} on top of SuperPoint~\cite{detoneSuperPointSelfSupervisedInterest2018} features, and
3) detector-free matchers including Sparse-NCNet~\cite{roccoEfficientNeighbourhoodConsensus2020} and DRC-Net~\cite{liDualResolutionCorrespondenceNetworks2020}.
For local features, we extract a maximum of 2K features with which we extract mutual nearest neighbors as the final matches. 
For methods directly outputting matches, we restrict a maximum of 1K matches, same as LoFTR.
We use the default hyperparameters in the original implementations for all the baselines.

\PAR{Results.} Tab.~\ref{tab:hpatches} shows that LoFTR notably outperforms other baselines under all error thresholds by a significant margin. 
Specifically, the performance gap between LoFTR and other methods increases with a stricter correctness threshold. 
We attribute the top performance to the 
larger number of match candidates provided by the detector-free design and the global receptive field brought by the Transformer. 
Moreover, the coarse-to-fine module also contributes to the estimation accuracy by refining matches to a sub-pixel level.

\begin{table}[tb]
    \centering
    \scriptsize{
        \vspace{-1.1cm}
\resizebox{\textwidth}{!}{%
\setlength\tabcolsep{4.0pt}
\renewcommand{\arraystretch}{1.2}
\begin{tabular}{llcccc}
\toprule
\multicolumn{1}{c}{\multirow{2}{*}{Category}} & \multicolumn{1}{c}{\multirow{2}{*}{Method}} & \multicolumn{3}{c}{Homography est. AUC} & \multirow{2}{*}{\#matches} \\ \cline{3-5}
\multicolumn{1}{c}{}                          & \multicolumn{1}{c}{}                        &@3px          &@5px         &@10px         &                            \\ \hline
\multirow{4}{*}{Detector-based}               & D2Net~\cite{dusmanuD2NetTrainableCNN2019}+NN                                 & 23.2          & 35.9          & 53.6          &  0.2K                          \\
                                              & R2D2~\cite{revaudR2D2RepeatableReliable}+NN                                  & 50.6          & 63.9          & 76.8          &      0.5K                      \\
                                              & DISK~\cite{tyszkiewiczDISKLearningLocal2020}+NN                                  & 52.3          & 64.9          & 78.9          &     1.1K                       \\
                                              & SP~\cite{detoneSuperPointSelfSupervisedInterest2018}+SuperGlue~\cite{sarlinSuperGlueLearningFeature2019}                      & 53.9              & 68.3              & 81.7              & 0.6K                           \\ \hline
\multirow{3}{*}{Detector-free}                & Sparse-NCNet~\cite{roccoEfficientNeighbourhoodConsensus2020}                                & 48.9          & 54.2          & 67.1          &  1.0K                          \\
                                              & DRC-Net~\cite{liDualResolutionCorrespondenceNetworks2020}                                     & 50.6          & 56.2          & 68.3          &   1.0K                         \\
                                              & \b{\shortname-DS}                   & \b{65.9}          & \b{75.6}          & \b{84.6}         &   1.0K                          \\ 
\bottomrule
\end{tabular}
}
    }
    \caption{\b{Homography estimation on HPatches~\cite{daiScanNetRichlyannotated3D2017}.} The AUC of the corner error in percentage is reported.
    The suffix DS indicates the differentiable matching with dual-softmax.}
    \label{tab:hpatches}
\end{table}

\subsection{Relative Pose Estimation}
\begin{table}[tb]
    \centering
    \scriptsize{
        \vspace{-1.1cm}
\setlength\tabcolsep{4.0pt}
\begin{tabular}{clcccc}
    \toprule
    \multirow{2}{*}[-.4em]{Category} & \multicolumn{1}{c}{\multirow{2}{*}[-.4em]{Method}} & \multicolumn{3}{c}{Pose estimation AUC}  \\
    \cmidrule(lr){3-5}
                                               &                                    & @5\degree & @10\degree & @20\degree           \\
    \midrule
    \multirow{7}{1.5cm}[-.4em]{Detector-based} & ORB~\cite{rubleeORBEfficientAlternative2011}+GMS~\cite{bian2017gms}                            & \05.21    & 13.65      & 25.36     \\
                                               & D2-Net~\cite{dusmanuD2NetTrainableCNN2019}+NN                         & \05.25    & 14.53      & 27.96     \\
                                              & ContextDesc~\cite{luo2019contextdesc}+Ratio Test~\cite{loweDistinctiveImageFeatures2004}             & \06.64    & 15.01      & 25.75     \\
                                               & SP~\cite{detoneSuperPointSelfSupervisedInterest2018}+NN                     & \09.43    & 21.53      & 36.40     \\
                                               & SP~\cite{detoneSuperPointSelfSupervisedInterest2018}+PointCN~\cite{yi2018learning}                 & 11.40     & 25.47      & 41.41     \\
                                               & SP~\cite{detoneSuperPointSelfSupervisedInterest2018}+OANet~\cite{zhang2019oanet}                   & 11.76     & 26.90      & 43.85     \\
                                               & SP~\cite{detoneSuperPointSelfSupervisedInterest2018}+SuperGlue~\cite{sarlinSuperGlueLearningFeature2019}               & 16.16     & 33.81      & 51.84     \\
    \midrule
    \multirow{4}{1.5cm}{Detector-free}
                                               & \multicolumn{1}{l}{DRC-Net \dag~\cite{liDualResolutionCorrespondenceNetworks2020}}   & 7.69      & 17.93       & 30.49      \\
                                               & \multicolumn{1}{l}{\shortname-OT\dag} & 16.88     & 33.62      & 50.62     \\
                                               & \multicolumn{1}{l}{\b{\shortname-OT}} & 21.51 & 40.39  & \b{57.96} \\
                                               & \multicolumn{1}{l}{\b{\shortname-DS}} & \b{22.06} & \b{40.8}  & 57.62 \\
    \bottomrule
\end{tabular}

    }
    \caption{\b{Evaluation on ScanNet~\cite{daiScanNetRichlyannotated3D2017} for indoor pose estimation.} The AUC of the pose error 
    in percentage is reported. LoFTR improves the state-of-the-art methods by a large margin. \dag indicates models trained on MegaDepth. 
    The suffixes OT and DS indicate differentiable matching with optimal transport and dual-softmax, respectively.}
    \label{tab:scannet}
\end{table}

\begin{table}[tb]
    \centering
    \scriptsize{

\setlength\tabcolsep{4.0pt}
\begin{tabular}{clcccc}
    \toprule
    \multirow{2}{*}[-.4em]{Category} & \multicolumn{1}{c}{\multirow{2}{*}[-.4em]{Method}} & \multicolumn{3}{c}{Pose estimation AUC}  \\
    \cmidrule(lr){3-5}
                                               &                                    & @5\degree & @10\degree & @20\degree  \\
    \midrule
    \multirow{1}{1.5cm}[-.0em]{Detector-based} & SP~\cite{detoneSuperPointSelfSupervisedInterest2018}+SuperGlue~\cite{sarlinSuperGlueLearningFeature2019}               & 42.18     & 61.16      & 75.96       \\
    \midrule
    \multirow{3}{1.5cm}{Detector-free}
                                               & \multicolumn{1}{l}{DRC-Net~\cite{liDualResolutionCorrespondenceNetworks2020}}        & 27.01     & 42.96      & 58.31       \\
                                               & \multicolumn{1}{l}{\b{\shortname-OT}} & 50.31 & 67.14  & 79.93   \\
                                               &
                                    \multicolumn{1}{l}{\b{\shortname-DS}} & \b{52.8} & \b{69.19}  & \b{81.18}   \\
    \bottomrule
\end{tabular}
    }
    \caption{\b{Evaluation on MegaDepth~\cite{liMegaDepthLearningSingleView2018} for outdoor pose estimation.} Matching with LoFTR results in better performance in the outdoor pose estimation task.}
    \label{tab:megadepth}
\end{table}

\PAR{Datasets.} We use ScanNet~\cite{daiScanNetRichlyannotated3D2017} and MegaDepth~\cite{liMegaDepthLearningSingleView2018} to demonstrate the effectiveness of LoFTR for pose estimation in indoor and outdoor scenes, respectively.

ScanNet contains 1613 monocular sequences with ground truth poses and depth maps. Following the procedure from SuperGlue~\cite{sarlinSuperGlueLearningFeature2019}, we sample 230M image pairs for training, with overlap scores between 0.4 and 0.8. We evaluate our method on the 1500 testing pairs from ~\cite{sarlinSuperGlueLearningFeature2019}. All images and depth maps are resized to $640 \times 480$. This dataset contains image pairs with wide baselines and extensive texture-less regions.

MegaDepth consists of 1M internet images of 196 different outdoor scenes. The authors also provide sparse reconstruction from COLMAP~\cite{schonbergerStructurefromMotionRevisited2016} and depth maps computed from multi-view stereo. We follow DISK~\cite{tyszkiewiczDISKLearningLocal2020} to only use the scenes of ``Sacre Coeur'' and ``St. Peter's Square'' for validation, from which we sample 1500 pairs for a fair comparison. Images are resized such that their longer dimensions are equal to 840 for training and 1200 for validation. 
The key challenge on MegaDepth is matching under extreme viewpoint changes and repetitive patterns.

\PAR{Evaluation protocol.} 
Following ~\cite{sarlinSuperGlueLearningFeature2019}, we report the AUC of the pose error at thresholds ($5^\circ,10^\circ,20^\circ$), where the pose error is defined as the maximum of angular error in rotation and translation. 
To recover the camera pose, we solve the essential matrix from predicted matches with RANSAC. 
We don't compare the matching precisions between LoFTR and other detector-based methods due to the lack of a well-defined metric (e.g., matching score or recall~\cite{heinlyComparativeEvaluationBinary2012, mikolajczykPerformanceEvaluationLocal2005b}) for detector-free image matching methods.
We consider DRC-Net~\cite{liDualResolutionCorrespondenceNetworks2020} as the state-of-the-art method in detector-free approaches~\cite{roccoNeighbourhoodConsensusNetworks2018,roccoEfficientNeighbourhoodConsensus2020}.

\begin{figure*}[btp]
\vspace{-1.5cm}
    \centering
    \includegraphics[width=0.8\linewidth]{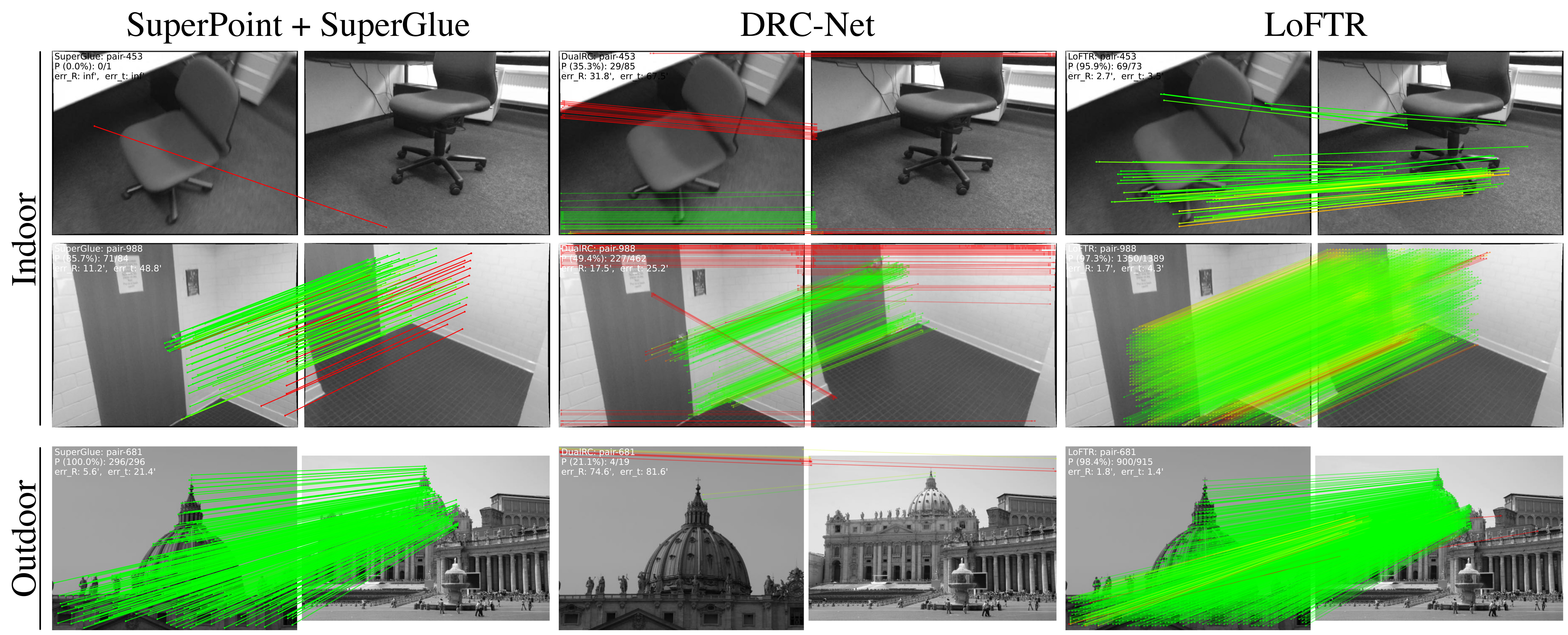}%
    \caption{{\textbf{Qualitative results.}} LoFTR is compared to SuperGlue~\cite{sarlinSuperGlueLearningFeature2019} and DRC-Net~\cite{liDualResolutionCorrespondenceNetworks2020} in indoor and outdoor environments. LoFTR obtains more correct matches and fewer mismatches, successfully coping with low-texture regions and large viewpoint and illumination changes. 
    The red color indicates epipolar error beyond $5 \times 10^{-4}$ for indoor scenes and $1 \times 10^{-4}$ for outdoor scenes (in the normalized image coordinates).
    More qualitative results can be found on the \href[pdfnewwindow]{https://zju3dv.github.io/loftr/}{project webpage}.
    }
    \label{fig:qua-allmatch}
\end{figure*}

\PAR{Results of indoor pose estimation.} LoFTR achieves the best performance in pose accuracy compared to all competitors (see Tab.~\ref{tab:scannet} and Fig.~\ref{fig:qua-allmatch}). %
Pairing LoFTR with optimal transport or dual-softmax as the differentiable matching layer achieves comparable performance.
Since the released model of DRC-Net$\dag$ is trained on MegaDepth, we provide the results of LoFTR$\dag$ trained on MegaDepth for a fair comparison.
LoFTR$\dag$ also outperforms DRC-Net$\dag$ by a large margin in this evaluation (see Fig.~\ref{fig:qua-allmatch}), which demonstrates the generalizability of our model across datasets.
\begin{table}[tb]
\vspace{-1.1cm}
    \centering
    \scriptsize{
        \resizebox{\textwidth}{!}{%
    \setlength\tabcolsep{4.0pt}
    \renewcommand{\arraystretch}{1.2}
    \begin{tabular}{lcc}
        \toprule
        \multirow{2}{*}{Method}                                                                                                          & Day                                                                       & Night                                  \\
        \cline{2-3}
                                                                                                                                         & \multicolumn{2}{c}{(0.25m,2\degree) / (0.5m,5\degree) / (1.0m,10\degree)}                                          \\ \hline

        \multicolumn{3}{l}{\textbf{Local Feature Evaluation on Night-time Queries}}                                                                                                                                                                           \\ \hline
        R2D2~\cite{revaudR2D2RepeatableReliable}+NN                                                                                      & -                                                                         & 71.2 / 86.9 / 98.9                     \\
        LISRD~\cite{Pautrat_2020lisrd}+SP~\cite{detoneSuperPointSelfSupervisedInterest2018}+AdaLam~\cite{cavalli2020handcrafted} & -                                                                         & \textbf{73.3} / 86.9 / 97.9            \\
        ISRF~\cite{melekhov2020isrf}+NN                                                                                                  & -                                                                         & 69.1 / 87.4 / 98.4                     \\
        SP~\cite{detoneSuperPointSelfSupervisedInterest2018}+SuperGlue~\cite{sarlinSuperGlueLearningFeature2019}                 & -                                                                         & \textbf{73.3} / 88.0 / 98.4            \\
        \b{LoFTR-DS}                                                                                                                         & -                                                                         & 72.8 / \textbf{88.5} / \textbf{99.0}   \\ \hline
        \multicolumn{3}{l}{\textbf{Full Visual Localization with HLoc}}                                                                                                                                                                                              \\ \hline
        SP~\cite{detoneSuperPointSelfSupervisedInterest2018}+SuperGlue~\cite{sarlinSuperGlueLearningFeature2019}                 & \textbf{89.8} / \textbf{96.1} / \textbf{99.4}                             & 77.0 / \textbf{90.6} / \textbf{100.0}  \\
        \b{LoFTR-OT}                                                                                                                         & 88.7 / 95.6 / 99.0                                                        & \textbf{78.5} / \textbf{90.6} / \099.0 \\ \bottomrule
    \end{tabular}%
}
    }
    \caption{\b{Visual localization evaluation on the Aachen Day-Night~\cite{zhangReferencePoseGeneration2020b} benchmark v1.1}. The evaluation results on both the local feature evaluation track and the full visual localization track are reported.}
    \label{tab:aachen_v1-1}
\end{table}

\PAR{Results of Outdoor Pose Estimation.}
As shown in Tab.~\ref{tab:megadepth},
LoFTR outperforms the detector-free method DRC-Net by 61\% at AUC@10\degree, demonstrating the effectiveness of the Transformer.
For SuperGlue, we use the setup from the open-sourced localization toolbox HLoc~\cite{sarlinCoarseFineRobust2019b}. 
LoFTR outperforms SuperGlue by a large margin (13\% at AUC@10\degree), which demonstrates the effectiveness of the detector-free design.
Different from indoor scenes, LoFTR-DS performs better than LoFTR-OT on MegaDepth. 
More qualitative results can be found in Fig.~\ref{fig:qua-allmatch}.

\subsection{Visual Localization}

\vspace{-0.1cm}

\PAR{Visual Localization.} 
Besides achieving competitive performance for relative pose estimation, LoFTR can also benefit visual localization, which is the task to estimate the 6-DoF poses of given images with respect to the corresponding 3D scene model. 
We evaluate LoFTR on the Long-Term Visual Localization Benchmark
~\cite{toftLongTermVisualLocalization2020}~(referred to as VisLoc benchmark in the following). 
It focuses on benchmarking visual localization methods under varying conditions, e.g., day-night changes, scene geometry changes, and indoor scenes with plenty of texture-less areas.
Thus, the visual localization task relies on highly robust image matching methods. 

\PAR{Evaluation.} 
We evaluate LoFTR on two tracks of VisLoc that consist of several challenges. 
First, the ``visual localization for handheld devices'' track requires a full localization pipeline.
It benchmarks on two datasets, the Aachen-Day-Night dataset~\cite{sattlerImageRetrievalImagebased2012, zhangReferencePoseGeneration2020b} concerning outdoor scenes and the InLoc~\cite{tairaInLocIndoorVisual2018} dataset concerning indoor scenes. 
We use open-sourced localization pipeline HLoc~\cite{sarlinCoarseFineRobust2019b} with the matches extracted by LoFTR. 
Second, the ``local features for long-term localization'' track provides a fixed localization pipeline to evaluate the local feature extractors themselves and optionally the matchers. 
This track uses the Aachen v1.1 dataset~\cite{zhangReferencePoseGeneration2020b}. 
We provide the implementation details of testing LoFTR on VisLoc in the supplementary material.

\begin{table}[tb]
\vspace{-1.1cm}
    \centering
    \scriptsize{
        \resizebox{\textwidth}{!}{%
    \setlength\tabcolsep{4.0pt}
    \renewcommand{\arraystretch}{1.2}
    \begin{tabular}{lcc}
        \toprule
        \multirow{2}{*}{Method}                                                                                                                                  & DUC1                                                                  & DUC2                                 \\
        \cline{2-3}
                                                                                                                                                                 & \multicolumn{2}{c}{(0.25m,10\degree) / (0.5m,10\degree) / (1.0m,10\degree)}                                        \\ \hline
        ISRF~\cite{melekhov2020isrf}                                                                                                                             & 39.4 / 58.1 / 70.2                                                    & 41.2 / 61.1 / 69.5                  \\
        KAPTURE~\cite{humenberger2020robust}+R2D2~\cite{revaudR2D2RepeatableReliable}                                                                            & 41.4 / 60.1 / 73.7                                                    & 47.3 / 67.2 / 73.3                   \\
        HLoc~\cite{sarlinCoarseFineRobust2019b}+SP~\cite{detoneSuperPointSelfSupervisedInterest2018}+SuperGlue~\cite{sarlinSuperGlueLearningFeature2019} & \textbf{49.0} / 68.7 / 80.8                                           & 53.4 / \textbf{77.1} / 82.4          \\
        HLoc~\cite{sarlinCoarseFineRobust2019b}+\b{LoFTR-OT}                                                                                                         & 47.5 / \textbf{72.2} / \textbf{84.8}                                  & \textbf{54.2} / 74.8 / \textbf{85.5} \\ \bottomrule
    \end{tabular}%
}
    }
    \addtolength{\textfloatsep}{-0.2in}
    \caption{\b{Visual localization evaluation on the InLoc~\cite{tairaInLocIndoorVisual2018} benchmark}.}
    \label{tab:inloc}
\end{table}

\begin{table}[tb]
    \centering
    \scriptsize{

\resizebox{\textwidth}{!}{%
    \setlength\tabcolsep{4.0pt}
    \begin{tabular}{lcccc}
        \toprule
        \multirow{2}{*}[-.4em]{Method}                                          & \multicolumn{3}{c}{Pose estimation AUC}                             \\
        \cmidrule(lr){2-4}
                                                                                & @5\degree                               & @10\degree & @20\degree & \\
        \midrule
        1) replace LoFTR with convolution                                       & 14.98                                   & 32.04      & 49.92        \\
        2) \nicefrac{1}{16} coarse-resolution + \nicefrac{1}{4} fine-resolution & 16.75                                   & 34.82      & 54.0         \\
        3) positional encoding per layer                                        & 18.02                                   & 35.64      & 52.77        \\
        4) larger model with $N_c=8, N_f=2$                                     & \b{20.87}                               & 40.23      & 57.56        \\
        \b{Full ($N_c=4, N_f=1$)}                                               & 20.06                                   & \b{40.8}   & \b{57.62}
        \\
        \bottomrule
    \end{tabular}
}

    }
    \caption{\b{Ablation study.} Five variants of LoFTR are trained and evaluated both on the ScanNet dataset.}
    \label{tab:ablation}
\end{table}

\PAR{Results.} We provide evaluation results of LoFTR in Tab.~\ref{tab:aachen_v1-1} and Tab.~\ref{tab:inloc}. 
We have evaluated LoFTR pairing with either the optimal transport layer or the dual-softmax operator and report the one with better results. 
LoFTR-DS outperforms all baselines in the local feature challenge track, showing its robustness under day-night changes. 
Then, for the visual localization for handheld devices track, 
LoFTR-OT outperforms all published methods on the challenging InLoc dataset, which contains extensive appearance changes, more texture-less areas, symmetric and repetitive elements.
We attribute the prominence to the use of the Transformer and the optimal transport layer, taking advantage of global information and jointly bringing global consensus into the final matches. 
The detector-free design also plays a critical role, preventing the repeatability problem of detector-based methods in low-texture regions. 
LoFTR-OT performs on par with the state-of-the-art method SuperPoint + SuperGlue on night queries of the Aachen v1.1 dataset and slightly worse on the day queries. 

\subsection{Understanding LoFTR}

\vspace{-.2cm}

\begin{figure*}[btp]
    \centering
    \includegraphics[width=0.85\linewidth]{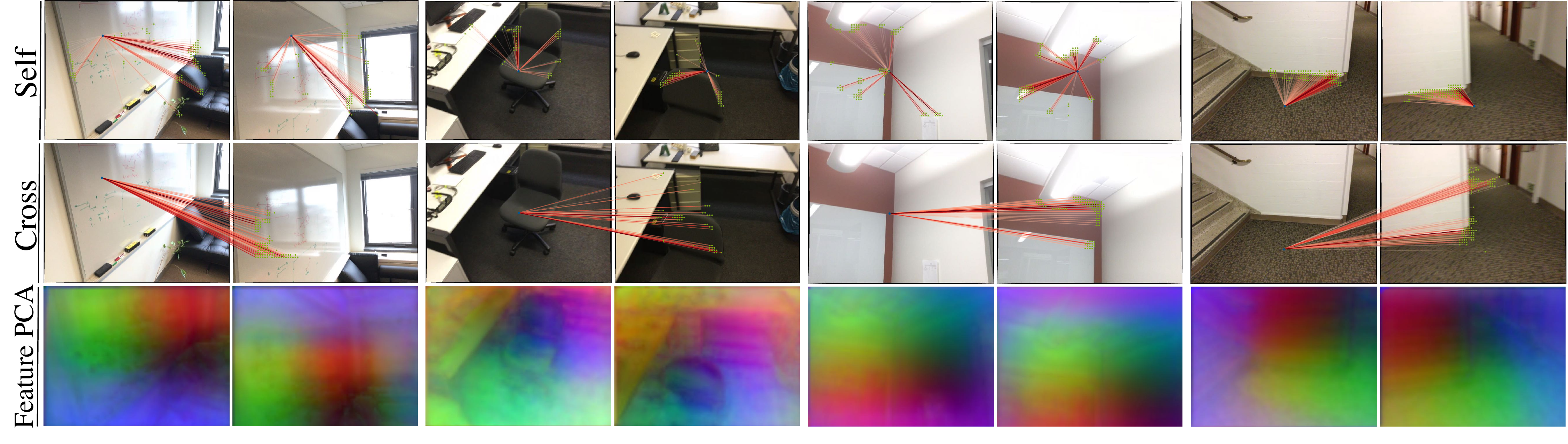}%
    \caption{{\textbf{Visualization of self and cross attention weights and the transformed features.}}  
    In the first two examples, the query point from the low-texture region is able to aggregate the surrounding global information flexibly. 
    For instance, the point on the chair is looking at the edge of the chair. 
    In the last two examples, the query point from the distinctive region can also utilize the richer information from other regions.
    The feature visualization with PCA further shows that LoFTR learns a position-dependent feature representation.
    }
    \label{fig:qua-attention}
\end{figure*}

\PAR{Ablation Study.} 
To fully understand the different modules in LoFTR, we evaluate five different variants with results shown in Tab.~\ref{tab:ablation}: 
1) Replacing the LoFTR module by convolution with a comparable number of parameters results in a significant drop in AUC as expected.
2) Using a smaller version of LoFTR with \nicefrac{1}{16} and \nicefrac{1}{4} resolution feature maps at the coarse and fine level, respectively, results in a running time of 104 ms and a degraded pose estimation accuracy.
3) Using DETR-style~\cite{carionEndtoEndObjectDetection2020} Transformer architecture which has positional encoding at each layer, leads to a noticeably declined result.
4) Increasing the model capacity by doubling the number of LoFTR layers to $N_c=8 \text{ and } N_f=2$ barely changes the results.
We conduct these experiments using the same training and evaluation protocol as indoor pose estimation on ScanNet with an optimal transport layer for matching.

\PAR{Visualizing Attention.} We visualize the attention weights in Fig.~\ref{fig:qua-attention}. 
 
\vspace{-0.1cm}
\section{Conclusion}
\vspace{-0.1cm}
This paper presents a novel detector-free matching approach, named \shortname, that can establish accurate semi-dense matches with Transformers in a coarse-to-fine manner. 
The proposed LoFTR module uses the self and cross attention layers in Transformers to transform the local features to be context- and position-dependent, which is crucial for LoFTR to obtain high-quality matches on indistinctive regions with low-texture or repetitive patterns.
Our experiments show that \shortname achieves state-of-the-art performances on relative pose estimation and visual localization on multiple datasets. 
We believe that LoFTR provides a new direction for detector-free methods in local image feature matching and can be extended to more challenging scenarios, e.g., matching images with severe seasonal changes.

\vspace{-0.1cm}
\PAR{Acknowledgement.}
The authors would like to acknowledge the support from the National Key Research and Development Program of China (No.~2020AAA0108901), NSFC (No.~61806176), and ZJU-SenseTime Joint Lab of 3D Vision.

\clearpage

{\small
\bibliographystyle{ieeefullname}
\bibliography{fulldb_cleaned_noarxiv}
}

\end{document}